# IndicRAGSuite: Large-Scale Datasets and a Benchmark for Indian Language RAG Systems

**Pasunuti Prasanjith**[1]    **Prathmesh B More**[1,2]
**Anoop Kunchukuttan**[1,2,3]    **Raj Dabre**[1,2]
[1]Nilekani Centre at AI4Bharat,
[2]Indian Institute of Technology Madras, India
[3]Microsoft, India

## Abstract

Retrieval-Augmented Generation (RAG) systems enable language models to access relevant information and generate accurate, well-grounded, and contextually informed responses. However, for Indian languages, the development of high-quality RAG systems is hindered by the lack of two critical resources: (1) evaluation benchmarks for retrieval and generation tasks, and (2) large-scale training datasets for multilingual retrieval. Most existing benchmarks and datasets are centered around English or high-resource languages, making it difficult to extend RAG capabilities to the diverse linguistic landscape of India. To address the lack of evaluation benchmarks, we create IndicMSMarco, a multilingual benchmark for evaluating retrieval quality and response generation in 13 Indian languages, created via manual translation of 1000 diverse queries from MS MARCO-dev set. To address the need for training data, we build a large-scale dataset of (question, answer, relevant passage) tuples derived from the Wikipedias of 19 Indian languages using state-of-the-art LLMs. Additionally, we include translated versions of the original MS MARCO dataset to further enrich the training data and ensure alignment with real-world information-seeking tasks. Resources are available here.

| Dataset | #Langs | Source | Size |
|---|---|---|---|
| NQ | 1 | Wiki | 307K |
| TriviaQA | 1 | Web Docs | 650K |
| SQuAD v1.1 | 1 | Wiki | 100K |
| MS MARCO | 1 | Web Docs | 8.8M |
| TREC-DL | 1 | Web Docs | 367K |
| MKQA | 26 | Wiki | 260K |
| TyDi QA | 11 | Wiki | 204K |
| BEIR | 1 | Diverse | Varies |
| **IndicRAGSuite** | **19** | **Wiki + MS MARCO** | **26M** |

Table 1: Statistics of existing retrieval model training datasets.

## 1 Introduction

Dense retrieval models have significantly advanced the field of information retrieval (IR), surpassing traditional methods such as BM25 in ad-hoc search and question-answering tasks. These models leverage dense vector representations to capture semantic relationships, enabling efficient retrieval of relevant documents through approximate nearest neighbor search. Such capabilities are pivotal for applications including web search, semantic similarity tasks, and Retrieval-Augmented Generation (RAG) systems, where dense retrievers allow language models to access external knowledge efficiently.

However, the success of dense retrieval models critically depends on the quality and scale of available training data and evaluation benchmarks (Karpukhin et al., 2020). While large-scale datasets such as MS MARCO (Nguyen et al., 2016), Natural Questions (Kwiatkowski et al., 2019), SQuAD (Rajpurkar et al., 2016), TriviaQA (Joshi et al., 2017), and HotpotQA (Yang et al., 2018) have propelled significant progress in English, the development of robust dense retrieval systems for under-resourced languages—particularly Indian languages—remains severely constrained (Xiong et al., 2021). Despite the demonstrated sample efficiency of dense retrieval models (Qu et al., 2021), the scarcity of large-scale supervised datasets for Indian languages continues to be a major bottleneck (Bonifacio et al., 2021). For instance, English datasets often contain millions of question-answer pairs, whereas datasets for Indian languages are limited to mere thousands (Zhang et al., 2021), further challenged by limited digital presence, script diversity, and dialectal variations (Jose and Bhattacharyya, 2021).

Multilingual benchmarks such as MIRACL (Zhang et al., 2023), MKQA (Longpre et al., 2021), NeuCLIR (Lawrie et al., 2023), MLQA (Lewis et al., 2020), and XQuAD (Artetxe et al., 2020) have contributed significantly to advancing cross-lingual retrieval. However, they predominantly focus on high-resource languages, with limited representation for Indian languages (Ruder et al., 2021). Specialized domain-centric datasets such as BioASQ (Nentidis et al., 2023), FiQA (Angelidis et al., 2020), and SciFact (Lo et al., 2020) also lack substantial Indian language coverage (Chakraborty and Bhattacharyya, 2022). As a result, there remains a critical gap: without adequate benchmarks or training data, it is difficult to build, evaluate, and systematically improve retrieval systems for India's rich and diverse linguistic landscape (Joshi et al., 2020).

To address this gap, we focus on creating essential infrastructure for Indian language retrieval:

**Key Contributions**

- **Multilingual Benchmark for 13 Indina languages:** We manually translate a subset of the MS MARCO dataset into 13 Indian languages, creating a multilingual benchmark (IndicMSMarco) for retrieval and response generation evaluation. This addresses the absence of standardized evaluation datasets for Indian languages and enables fair, systematic comparisons.

- **Scalable Synthetic Dataset:** We construct a large-scale dataset comprising around 14 million (question, answer, relevant passage) triplets across 19 Indian languages. This dataset is generated by leveraging Wikipedia's multilingual content and large language models to create diverse and reasoning-rich examples. In addition, we translate the MS MARCO train and dev sets into 14 Indian languages, enabling supervised training of dense retrievers in a multilingual setup. Together, these datasets substantially expand the resources available for training retrieval models for Indian languages.

## 2 Related Work

### 2.1 Multilingual Benchmarks for Evaluation

Recent multilingual retrieval benchmarks offer valuable insights but remain inadequate for Indian languages. XOR-Retrieve (Asai et al., 2021) includes only Bengali and Telugu and focuses on English-centric retrieval, limiting its monolingual utility. MIRACL (Zhang et al., 2023) covers just three Indian languages and is restricted to Wikipedia, which lacks regional depth. XTREME-UP (Ruder et al., 2023), though aimed at low-resource settings, suffers from noisy task inclusion and struggles with script diversity. Common shortcomings across these efforts include limited language coverage, inconsistent evaluation, and translation artifacts—hindering the development of robust retrieval systems for India's linguistically diverse population.

To overcome these limitations, we introduce **IndicMSMARCO**, a multilingual retrieval benchmark tailored for Indian languages, adapting the high-quality MS MARCO framework (Nguyen et al., 2016) to regional contexts. Our benchmark comprises 1,000 diverse queries and passages from MS MARCO, spanning topics like science, history, and technology, with balanced complexity and length. Queries and passages are first translated into 13 Indian languages using LLaMA 3.3 70B (Research, 2024), and then manually verified and post-edited by expert annotators to ensure high-quality translation. This post-editing process ensures linguistic and semantic fidelity, addressing accuracy, fluency, and consistency, with particular care for named entities and cultural nuances. IndicMSMARCO supports monolingual retrieval, addresses script diversity, and provides standardized evaluation metrics—filling a critical gap in benchmarking retrieval systems for Indian languages.

### 2.2 Multilingual Retrievers and Training Data Requirements

Several multilingual retrieval models have emerged with varying architectures and capabilities. Early baselines like mBERT (Pires et al., 2019) and XLM-R (Conneau et al., 2020) focused on cross-lingual understanding via masked language modeling and have since been adapted for retrieval. mT5 (Xue et al., 2021) introduced a text-to-text paradigm and has been used in both dual-encoder and generative retrieval settings. Dense retrievers such as mDPR (Asai et al., 2021) and mContriever (Izacard et al., 2022) leveraged parallel data and contrastive learning, respectively, while mE5 (Wang et al., 2022) used multitask learning across 100+ languages to directly optimize retrieval performance. Proprietary systems like OpenAI's text-embedding-ada-002 (Neelakantan et al., 2022) and Voyage AI (Voyage AI, 2023) show strong multilingual performance, though their training remains opaque. More recently, jina-embeddings (Günther et al., 2024) target long-context retrieval but still trail closed-source models. Despite these advances, performance remains inconsistent across language fami-lies, underscoring the need for inclusive and task-specific multilingual training data.

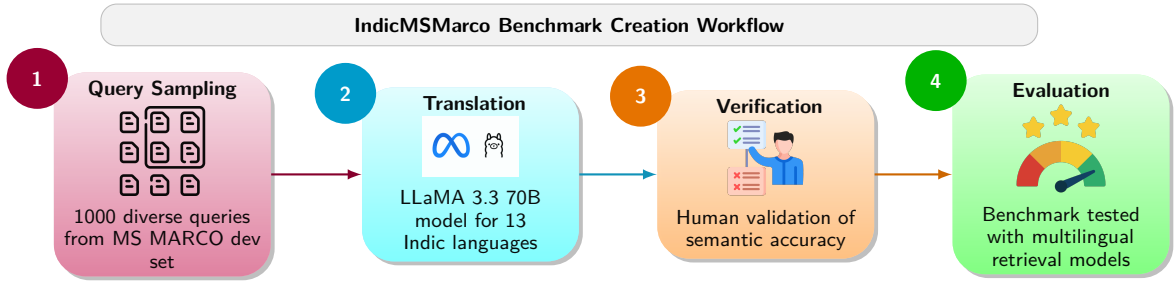

Figure 1: Benchmark creation workflow for IndicMSMarco: from query selection to human-verified multilingual evaluation.

The availability of high-quality training data remains a key bottleneck for multilingual retrieval systems. Existing resources typically rely on parallel corpora (e.g., Wikipedia translations in mDPR (Asai et al., 2021)), web-mined text pairs (e.g., mC4 in mContriever (Izacard et al., 2022)), and limited human-annotated datasets like mMARCO (Bonifacio et al., 2021). These datasets, however, are heavily skewed toward English—with MS MARCO offering 8.8M queries (Nguyen et al., 2016), while Indian languages have access to only a fraction of that volume. A recent effort to address these gaps is the INDIC-MARCO dataset (Haq et al., 2023), which translates MS MARCO into 11 Indian languages using NLLB-1.3B-Distilled via CTranslate2. However, its sentence-level translation strategy fragments context, potentially reducing semantic fidelity.

To address data limitations, we construct a large-scale multilingual training dataset using Wikipedia dumps from 19 Indian languages and generate question-answer-reasoning triplets via the Llama 3.3 70B model. Unlike prior approaches that split passages before translation, we retain full-paragraph structure and employ IndicTrans3-beta (AI4Bharat) to ensure semantic coherence. We also translate the MS MARCO training and dev sets into 14 Indian languages.

## 3 IndicMSMARCO Benchmark

To advance retrieval models for Indian languages, we introduce **IndicMSMARCO**, a multilingual retrieval benchmark. MS MARCO (Nguyen et al., 2016) is a large-scale dataset designed for question answering, passage ranking, and document retrieval tasks. It comprises real-world queries from Bing search logs, with relevant passages annotated by human assessors. While MS MARCO has served as a cornerstone for retrieval research in English, the absence of comparable high-quality benchmarks for Indian languages has hindered the development of robust retrieval systems in these languages.

To address this gap, we adapt MS MARCO by creating a multilingual variant specifically tailored for Indian languages. Our benchmark consists of 1,000 carefully selected queries and their corresponding passages from the MS MARCO development set. The selection process prioritizes:

- **Topic Diversity:** Ensuring a wide range of subject areas, including science, history, politics, health, and technology.

- **Query Complexity Variation:** Incorporating simple factual queries, descriptive queries, and complex entity-based queries.

- **Balanced Representation:** Ensuring a mix of short, medium, and long-form queries to evaluate retrieval models comprehensively.

We construct the IndicMSMARCO benchmark in two phases: (1) automatic translation of queries and passages using the Llama 3.3 70B model, and (2) human verification, correction, and annotation to ensure linguistic and semantic fidelity. An illustrative example of a Hindi query–answer–passage triplet from IndicMSMARCO is shown in Figure 2.

### 3.1 Automated Translation with Llama 3.3 70B

To generate high-quality multilingual versions of MS MARCO queries and passages, we leverage the Llama 3.3 70B model, a state-of-the-art generative language model with strong multilingual capabilities. The translation pipeline follows a structured approach:

- **Query Translation:** Each query from the selected MS MARCO subset is translated into

| | |
|---|---|
| **Examples of query–answer–passage triplet in Hindi (hi) from IndicMSMarco.** | |
| **Query:** | कौन सा रक्त प्रकार सबसे अधिक बार होता है |
| **Answer:** | ओ पॉजिटिव |
| **Passage:** | रक्त प्रकार और जनसंख्या। ओ पॉजिटिव सबसे आम रक्त प्रकार है। सभी जातीय समूहों में इन रक्त प्रकारों का समान मिश्रण नहीं होता है। उदाहरण के लिए, हिस्पैनिक लोगों में ओ रक्त प्रकार की संख्या अपेक्षाकृत अधिक होती है, जबकि एशियाई लोगों में बी रक्त प्रकार की संख्या अपेक्षाकृत अधिक होती है। यू.एस. जनसंख्या में विभिन्न रक्त प्रकारों का मिश्रण इस प्रकार है: |

Figure 2: Benchmark Example in Hindi

13 major Indian languages. Llama 3.3 70B ensures the retention of query intent while adapting language-specific structures.

- **Passage Translation:** The corresponding passages are translated using context-aware generation, ensuring coherence and fidelity to the original English passage. The model is prompted to preserve named entities, numerical data, and domain-specific terminology to maintain retrieval relevance.

The automated translation process enables rapid expansion of the benchmark to multiple Indian languages. However, machine translations often introduce errors related to syntax, semantic drift, and ambiguity. To ensure quality, we conduct a rigorous human verification and annotation phase.

### 3.2 Human Verification and Annotation

After translating queries and passages into multiple languages through LLaMA 3.3 70B, we employ a structured human annotation process to validate, correct, and refine translations. This phase involves expert linguists, native speakers, and bilingual annotators across different Indian languages.

The verification process follows three key steps:

- **Linguistic Accuracy Check:** Annotators review translations for grammatical correctness, fluency, and readability. This step ensures that the translated queries and passages adhere to the natural syntax and style of each language.

- **Semantic Consistency Evaluation:** Each query and passage pair is cross-checked against the original English version to verify that the meaning remains intact. Annotators flag and correct any instances of semantic drift, mistranslations, or ambiguous phrasing.

- **Entity and Domain-Specific Validation:** To maintain retrieval relevance, experts validate technical terms, named entities (e.g., locations, person names, numerical values), and context-sensitive information. Necessary corrections are made to preserve factual and contextual accuracy.

In addition to validation, annotators actively correct translation errors to ensure precision and naturalness in every language. This meticulous verification and correction process ensures that IndicMSMARCO serves as a high-quality, reliable benchmark for evaluating retrieval models in Indian languages. By incorporating both automated translation and human refinement, we create a dataset that is not only scalable but also linguistically robust.

### 3.3 Significance of IndicMSMARCO

The IndicMSMARCO benchmark is a crucial resource for the development of dense retrieval models tailored to Indian languages. It enables:

- **Standardized Evaluation:** Providing a common ground for comparing retrieval performance across multiple Indian languages.

- **Enhanced Multilingual Retrieval Research:** Facilitating the training and fine-tuning of retrieval models for underrepresented languages.

- **Real-World Applicability:** Addressing practical challenges in multilingual search systems, digital libraries, and knowledge retrieval applications in India.

By constructing IndicMSMARCO, we take a significant step toward bridging the linguistic gap in information retrieval and fostering equitable access to advanced retrieval technologies across diverse Indian languages.

| Language | Multilingual e5-small | Multilingual e5-base | Multilingual e5-large | LLM2VEC LLaMA 3.1 8B Instruct | BGE-M3 |
| --- | --- | --- | --- | --- | --- |
| Assamese | 0.30 | 0.40 | 0.45 | 0.42 | **0.46** |
| Bengali | 0.39 | 0.46 | 0.48 | 0.44 | **0.49** |
| Gujarati | 0.34 | 0.43 | **0.48** | 0.42 | 0.48 |
| Hindi | 0.44 | 0.49 | **0.52** | 0.49 | 0.52 |
| Kannada | 0.38 | 0.44 | **0.47** | 0.40 | 0.47 |
| Malayalam | 0.38 | 0.45 | 0.49 | 0.43 | **0.49** |
| Marathi | 0.36 | 0.45 | **0.49** | 0.45 | 0.49 |
| Nepali | 0.39 | 0.45 | 0.49 | 0.45 | **0.49** |
| Odia | 0.31 | 0.39 | 0.45 | 0.34 | **0.45** |
| Punjabi | 0.32 | 0.42 | **0.48** | 0.42 | 0.48 |
| Tamil | 0.38 | 0.45 | **0.49** | 0.40 | 0.49 |
| Telugu | 0.39 | 0.45 | 0.50 | 0.42 | **0.50** |
| Urdu | 0.35 | 0.45 | **0.49** | 0.44 | 0.48 |

Table 2: MRR scores on IndicMSMarco Benchmark for 13 Indian languages using various dense retrieval models. Highest scores per language are in **bold**.

## 3.4 Experiments and Results

We evaluate the performance of various dense retriever models on IndicMSMarco Benchmark across 13 major Indian languages using **Mean Reciprocal Rank (MRR)** as the evaluation metric. The models compared include **LLM2VEC (LLaMA 3.1 8B Instruct)**, **BGE-M3**, and the **Multilingual E5** family—*e5-small*, *e5-base*, and *e5-large*. These models span multilingual, instruction-tuned, and retrieval-centric architectures, offering insights into their strengths and limitations in multilingual Indian language settings.

As shown in Table 2, **BGE-M3** achieves the best or near-best MRR in the majority of languages, leading in 8 out of 13 languages. Notably, it scores **0.49** in Malayalam and Tamil, and **0.50** in Telugu, indicating strong generalization across linguistically diverse Indian scripts.

**Multilingual e5-large** also performs consistently well, obtaining the highest score in 4 languages, including **Hindi (0.52)**, **Gujarati (0.48)**, and **Urdu (0.49)**. The steady improvement from e5-small to e5-large demonstrates the benefits of scaling for multilingual retrieval effectiveness. The smaller e5 models still deliver respectable performance, particularly in medium-resource languages.

**LLM2VEC**, based on the LLaMA 3.1 8B architecture and fine-tuned for retrieval tasks, shows competitive results across several languages. For example, it achieves **0.49** in Hindi and Marathi, and **0.45** in Nepali. While it does not dominate across all languages, its results show that instruction-tuned LLMs are viable alternatives for dense retrieval in multilingual contexts.

Languages such as **Hindi** consistently receive high MRR scores across all models, likely due to better representation in training corpora. In contrast, **Assamese** and **Odia** score lower overall, reflecting the challenges of retrieval in lower-resource languages.

Overall, the benchmark results in Table 2 highlight the strength of modern dense retrievers like **BGE-M3** and **Multilingual e5-large**, particularly in multilingual and low-resource settings. These findings establish a strong baseline for Indian language retrieval and point toward key directions for future work on multilingual and underrepresented language support.

## 4 RAG Training Dataset Construction

The effectiveness of any information retrieval (IR) system is largely dependent on the quantity and quality of its training data. Advancing IR research in Indian languages has historically been hindered by the scarcity of large-scale, high-quality datasets. To address this gap, we constructed two complementary training datasets: (1) a Wikipedia-generated multilingual dataset of question-answer-reasoning triplets, and (2) a high-quality translated version of the MS MARCO dataset adapted for Indian languages.

### 4.1 Wikipedia-Based Question-Answer-Reasoning Dataset

#### 4.1.1 Dataset Design and Objectives

Our primary objective was to construct a linguistically diverse dataset for training retriever models in Indian languages. The Wikipedia-based data set was designed to meet the following criteria.

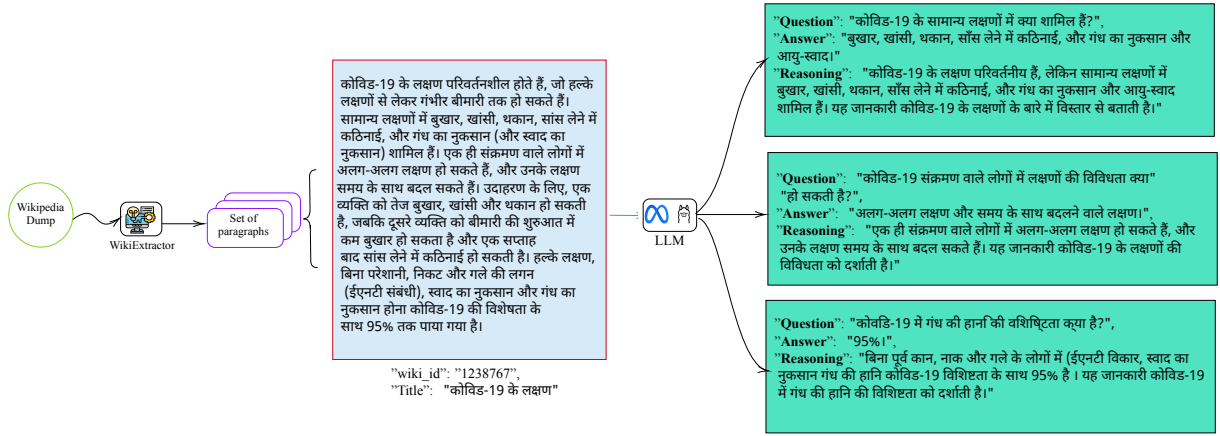

Figure 3: The data processing pipeline—from raw Wikipedia dumps to paragraph extraction and an LLM—generating Hindi Q&A pairs with explanatory reasoning

- **Scale:** Millions of question-answer-reasoning triplets per language to support robust model training.

- **Diversity:** Coverage across a wide range of topics and linguistic nuances reflecting India's cultural and regional diversity.

- **Quality:** Contextually accurate, grammatically correct, and semantically meaningful triplets.

- **Multilingual Coverage:** Broad applicability across 19 major Indian languages.

### 4.1.2 Source: Wikipedia Dumps

To construct the data set, we used Wikipedia dumps, compressed archives that contain full-text articles in various languages. Wikipedia serves as an ideal source due to the following:

- **Multilingual Availability:** Coverage across all 19 targeted Indian languages.

- **Topic Diversity:** Wide-ranging subject matter, including science, history, culture, and current events.

- **Open Access:** Unrestricted usage, allowing the creation of large-scale data sets.

### 4.1.3 Data Extraction and Preprocessing

We processed raw Wikipedia dumps using `WikiExtractor`, cleaning the extracted content through:

- Removal of metadata, HTML tags, formatting, and hyperlinks.

- Segmentation of articles into paragraph-level chunks to ground question-answer pairs in localized contexts.

Paragraph-level segmentation was crucial to ensure that the generated questions and answers maintained a tight contextual relevance.

### 4.1.4 Triplet Generation Using LLaMA 3.3 70B

To generate high-quality question-answer-reasoning triplets for Indian languages, we curated a pipeline that transforms raw Wikipedia content into structured QA data. As illustrated in Figure **??**, the process begins with extracting paragraphs from Wikipedia dumps using the `WikiExtractor` tool. Each extracted paragraph is associated with metadata such as the article title and a unique wiki ID.

We then use the LLaMA 3.3 70B model to generate structured triplets for each paragraph. Specifically, the model produces three distinct question-answer pairs, each accompanied by a detailed reasoning segment. This reasoning component is crucial—it ensures that the answer is grounded in the paragraph and that the model interprets content beyond superficial keyword matching. Moreover, the triplets are crafted to cover diverse question types (e.g., "what," "why," "how," "when") and different parts of the paragraph, thereby reducing bias toward the initial lines.

Key aspects of this step include:

- **Comprehensiveness:** Questions are generated to span the full semantic content of the paragraph, promoting diverse information coverage.

- **Reasoning-driven generation:** The addition of explanatory reasoning promotes deeper understanding and better supports answer validity.

- **Multilingual robustness:** The LLaMA 3.3 70B model was prompted to adhere to the grammatical and syntactic structures of each target Indian language.

Figure 3 demonstrates an example in Hindi. The paragraph describes COVID-19 symptoms, from which the model generates semantically varied questions. Each question is paired with an appropriate answer and a reasoning span that justifies the answer choice using explicit context from the paragraph.

### 4.1.5 Scale and Multilingual Coverage

The final dataset comprises approximately 14 million question-answer-reasoning triplets across 19 Indian languages. This large-scale dataset is designed to support robust training and evaluation of multilingual information retrieval (IR) models in linguistically diverse and low-resource settings.

To ensure the quality and utility of the dataset, we incorporated a filtering step as part of our data curation pipeline. During this stage, paragraphs that were either too short (lacking sufficient context) or excessively long (risking coherence issues or hallucination by the LLM) were excluded. This filtering was applied prior to triplet generation to maximize consistency and relevance in the resulting data.

Table 3 provides detailed statistics for each language, including the number of paragraphs and triplets both before and after filtering.

### 4.2 Translated MS MARCO Dataset

While the Wikipedia-based dataset offers wide topical diversity and supports paragraph-grounded reasoning, it lacks the structured, real-world query characteristics critical for training effective retrieval models. To address this, we constructed a translated version of the MS MARCO dataset specifically tailored for Indian languages. Our translation pipeline begins by selecting queries and corresponding passages from the original MS MARCO training and development sets. These were translated into 14 Indian languages using IndicTrans3-beta, a state-of-the-art translation model fine-tuned for Indian language translation tasks. Unlike prior efforts such as IndicIRSuite(Haq et al., 2023), which translated sentence-level fragments after splitting passages using tools like Moses SentenceSplitter, our method preserves full-paragraph structure throughout translation. This approach maintains better contextual coherence, semantic alignment, and domain fidelity.

Table 3: Wikipedia Generated Training Data Statistics for Each Language

| Language | Before Filtering | After Filtering |
|---|---|---|
| Assamese | 333,705 | 217,018 |
| Bengali | 3,320,042 | 2,060,963 |
| English | 6,384,632 | 4,109,199 |
| Gujarati | 354,824 | 245,063 |
| Hindi | 2,220,115 | 1,182,023 |
| Kannada | 1,114,088 | 670,236 |
| Kashmiri | 29,487 | 1,138 |
| Maithili | 92,722 | 38,028 |
| Malayalam | 1,371,674 | 901,402 |
| Manipuri | 46,458 | 31,389 |
| Marathi | 200,000 | 96,820 |
| Nepali | 402,100 | 222,597 |
| Odia | 268,239 | 175,743 |
| Punjabi | 689,306 | 393,769 |
| Santali | 189,066 | 97,963 |
| Sindhi | 250,836 | 118,869 |
| Tamil | 946,544 | 507,664 |
| Telugu | 3,276,885 | 1,824,025 |
| Urdu | 199,999 | 27,575 |
| **Total** | **21,740,681** | **13,927,586** |

While Indic-MARCO employed the int-8 quantized version of the NLLB-1.3B Distilled model primarily for translation efficiency, we prioritized translation quality and linguistic richness, selecting IndicTrans3-beta(AI4Bharat) for its superior BLEU scores and fluency in Indian languages. Special attention was paid to preserving the original search intent in queries and minimizing distortions caused by automatic translation.

This high-fidelity, paragraph-level translated MS MARCO dataset enables more realistic, task-specific training of dense retrievers for Indian languages. It complements our Wikipedia-based dataset by adding real-world, query-driven examples, thus facilitating robust retrieval performance across both open-domain and structured query scenarios. Through deeper linguistic integrity, broader language coverage, and stronger alignment with the retrieval task, our approach provides a substantially improved training resource compared to previous multilingual adaptations of MS MARCO.

### 4.3 Future Work

With the construction of high-quality multilingual datasets—comprising a Wikipedia-based question-answer-reasoning corpus and translated version of MS MARCO—the next phase of our work will focus on training and evaluating dense retriever models using these resources. This includes fine-tuning already existing retrieval architectures to understand the individual and combined impact of synthetic data and real-world query-passage pairs. We aim to benchmark performance across 13 Indian languages, with special emphasis on gains in low-resource language settings. Additionally, future directions include integrating domain-specific corpora like legal or medical texts, and incorporating human-in-the-loop refinement, ultimately moving toward the development of robust, open-domain multilingual IR systems tailored for Indian language users.

Table 4: Translated MS Marco Training Data Statistics by Language

| Language | Code | # Train Dataset | # Val Dataset |
|---|---|---|---|
| Assamese | asm | 778,638 | 97,941 |
| Bengali | ben | 778,638 | 97,941 |
| Gujarati | guj | 778,638 | 97,941 |
| Hindi | hin | 778,638 | 97,941 |
| Kannada | kan | 778,638 | 97,941 |
| Malayalam | mal | 778,638 | 97,941 |
| Marathi | mar | 765,873 | 97,941 |
| Nepali | nep | 754,154 | 97,941 |
| Odia | ori | 782,282 | 97,941 |
| Punjabi | pan | 778,638 | 97,941 |
| Sanskrit | san | 778,638 | 97,941 |
| Tamil | tam | 778,638 | 97,941 |
| Telugu | tel | 778,638 | 97,941 |
| Urdu | urd | 770,089 | 97,941 |
| **Total** | | **10,848,130** | **1,371,174** |

## 5 Conclusion

We present **IndicMSMARCO**, a human-verified multilingual benchmark for information retrieval in 13 Indian languages. By adapting the MS MARCO development set using Llama 3.3 70B and expert linguistic correction, IndicMSMARCO maintains semantic accuracy and fluency across diverse queries and topics. It enables standardized evaluation of retrieval models in low-resource Indian language settings.

To support model training, we introduce a dual-source corpus that combines contextually translated MS MARCO data with a large-scale Wikipedia-based dataset. This hybrid strategy captures both real-world search relevance and broad domain knowledge, enhancing model generalization across diverse IR scenarios in Indian languages.

**Appendix A: Prompt Template for Question-Answer-Reasoning Generation from Wikipedia Articles**

**System Prompt:**
*You are a precise and helpful Question-Answer Generator that creates factual questions with verifiable answers from provided content in <target_language>.*

**Task Prompt:**
You will first be given an example of how the desired output will look like. Then you will be given the content based on which you have to generate up to three challenging, logically coherent questions that strictly meet the following criteria:

1. **Standalone & Additional Context-Independent:** The questions should be understandable without additional context and must not contain any references to "the paragraph" or "the article" outside of the content provided.

2. **Unambiguous Answer:** Each question should have a single, clear, and factual answer.

3. **Grounded in Context & Conceptual Format:** Each question must be conceptually rooted in the provided article's content and follow this format:

    - Start with a clear question word (e.g., *What*, *How*, *Where*, *When*).

    - Integrate key information from the article smoothly, using logical connectors (e.g., "in relation to", "compared to", "as a result of", "which also", "in addition to").

    - If no valid questions can be generated from the content, do not generate any questions.

For each question:
- Provide the answer in parentheses after the question. The answer can be either one word or a phrase.
- Clearly explain the reasoning process, using an excerpt from the article as a reference.
- Do not use mixed language for numbering; always use the format "Question 1", "Question 2", etc. Avoid non-English numbering even for non-English datasets.
- Except for numbering headers, the questions, answers, and reasonings should be in the same language as the article, which is <target_language>.

**Example:**

   **Question 1:** [Sample question]
   **Reasoning:** [Explanation referencing article content]

**Content:** [Title]: [Article Text]

Figure 4: System and task prompt used for generating high-quality, language-specific question-answer pairs from article content.